# AI Agents in Emergency Response Applications

**Aryan Naim**
aryan.e.naim@jpl.nasa.gov

**Ryan Alimo**
sralimo@jpl.nasa.gov

**Jay Braun**
Jay.E.Braun@jpl.nasa.gov

*Abstract*—**Emergency personnel respond to various situations ranging from fire, medical, hazardous materials, industrial accidents, to natural disasters. Situations such as natural disasters or terrorist acts require a multifaceted response of firefighters, paramedics, hazmat teams, and other agencies. Engineering AI systems that aid emergency personnel proves to be a difficult system engineering problem. Mission-critical "edge AI" situations require low-latency, reliable analytics. To further add complexity, a high degree of model accuracy is required when lives are at stake, creating a need for the deployment of highly accurate, however computationally intensive models to resource-constrained devices. To address all these issues, we propose an agent-based architecture for deployment of AI agents via 5G service-based architecture.**

■ AFTER SEVERAL years of prototyping Artificial Intelligence (AI) algorithms exposed either via web services or deployed locally to mobile devices on top of existing 3G/4G networks in the United States and Canada, the authors have reached several preliminary conclusions. For example, a single monolithic application or even a suite of monolithic applications cannot satisfy the ever-changing situations that first responders must deal with during an emergency [1]. First responders are typically deployed to unpredictable environments and United States Department of Homeland Security (DHS) has requested artificial intelligent (AI) enabled technologies that can reduce the risk of injury via audio processing, natural language processing, chemicals detection, fire detection, route planning, and overall situational awareness. Many of the mentioned capabilities were prototyped via monolithic apps or several apps that together attempted to provide these capabilities. The authors' previous experiments exposed several major challenges.

OBSERVATIONS

Current AI software has an over-reliance on constant connectivity to backend services. Additionally, there is difficulty in the deployment of AI models to edge devices, in an ad-hoc manner with minimal reconfiguration. Furthermore, having onboard AI for near real-time multimedia analytics introduces immense strains for resource constrained devices[2]. We have observed overall mobile device performance degradation, and rapid draining of battery to 50% charge or less, after only 4 to 6 hours of operation, across several different Android mobile phones all running our AI application. A possible solution to all these issues may be the use of modular software that can be deployed as needed locally and remove itself when no longer in need, hence agent based computing[3,4].

### AI AGENTS

Agent-based computing is not a new endeavor; researchers and engineers have deployed swarms of AI agents that are inspired by nature[3,4]. AI agents used to troubleshoot telecommunications networks were studied extensively and deployed by BT Laboratories throughout the 1990s[5]. Backend AI services are only accessible if there are high-bandwidth and reliable networks nearby, and our previous field experiments demonstrated that 4G networks alone were not up to the task. Due to recent breakthroughs in the deployment of deep learning models to edge devices, users can take advantage of high accuracy multimedia analytics in mobile environments[6]. The combination of 5G's increased throughput and Service Based Architecture (SBA) may offer AI services either via web services or the ability to deploy agents at the edge. 5G's SBA offers discoverable services that any authorized consumer application can access[7]. Jet Propulsion Laboratory's (JPL) interest in AI agents stems from decades of onboard autonomy software research and deployment on interplanetary rovers[8,9]. Furthermore, JPL welcomes research that can modularize complex AI software, to ease the deployment of algorithms to remote platforms. In addition, the authors have been involved in the research and deployment of pervasive software, specifically AI and machine learning for emergency situations for several years. Additionally, due to recent and frequent wildfires, other working groups within JPL have researched technologies related to unmanned aerial systems for wildfire monitoring and management, such as JPL's BlueSky program[10].

### COMMUNICATION ARCHITECTURES

**Communication Architecture 1**

Before proposing a new architecture, we shall outline the typical way AI applications are integrated into enterprise applications. Using a typical client-server architecture, emergency personnel may send queries to 5G's service-based architecture (SBA) for AI capabilities. In response the server sends back available services. This exposure of services is identical to previous generations of service-oriented architectures (SOA) used in enterprise computing, which expose services via Extensible Markup Language (XML) or JavaScript Object Notation (JSON) to clients[11]. The querying of edge services may be initiated via many interfaces such as a call to special telephone numbers, or via apps that broadcast their location via Global Position System (GPS) and in return get a server response of nearby AI capabilities via 5G networks. Based on available services users can upload images to a URL or number for analysis. The multimedia is sent across 5G or other available networks to the analytics backend and knowledge is again returned via a message. This architecture is just a request and response service. Although client-server service-based architectures are dominant in many applications, they have little to no advantage when dealing with issues related to limited connectivity, low latency, high availability, and efficient use of limited on board resources. The disadvantages are similar to previous approaches where there is complete reliance on backend services. If a mobile device loses 5G or legacy connections, then the emergency personnel will lose all AI capability.

**Communication Architecture 2**

In the architecture proposed in **Figure 1**, in addition to data being exchanged for legacy support, actual AI agents will be deployed to edge devices such as mobile devices, wearable devices, smart cameras, or nearby small cell towers or all[11]. An AI agent differs from a monolithic program, since it performs specific analysis, and this single purpose program may be deployed for a limited amount of time with an expiration date[12]. In this model emergency personnel attempt to query 5G's service-based architecture (SBA) for AI capabilities, and instead of getting a list of available backend services, the user gets an equivalent list of AI agents that can be automatically deployed with a single click or even a voice command. Unlike the typical client-server architecture, a request for analysis does not exchange data, rather the agent management software deploys AI agents with embedded pre-trained models. The entire classification is performed onboard via a pre-trained model, and knowledge is returned to the user. The user only gets AI capabilities that are relevant to the emergency and based on current network conditions. The advantage of this local agent deployed architecture is significantly reduced reliance on backend services. There are even possibilities of collaborative AI agent utilizing 5G in conjunction with Peer to Peer (P2P) networks such as Bluetooth, for more resilient network connectivity solution with multiple networking options in case a certain network fails[13,14]. In a hypothetical edge AI model, an image processing agent and natural language processing (NLP) agent will analyze the situation on the ground, and automatically request other specific AI agents without any user interaction. Furthermore, the agents can clean up by uninstalling themselves, after the emergency is over, to spare on-board resources such as memory and battery power.

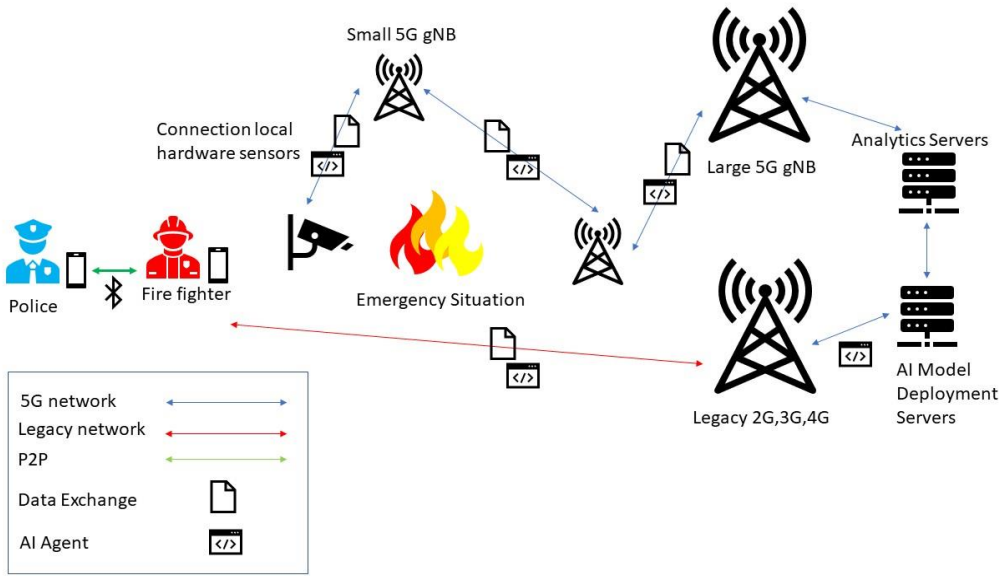

**Figure 1. Network based on both data exchange and AI agent deployment.**

## PREVIOUS FIELD EXPERIMENTS

### Localization

In situations where constant and accurate situational awareness can mean the difference between life and death, location-based analytics may be used to track the current position of firefighters as they rush in and out of a burning building. For instance, in field experiments with the cooperation of Grant County, Washington fire fighters, we observed GPS degradation that incorrectly mapped a person's location as outside of a building, while the person was in fact inside of the building (**Figure 2**).

We used two different mapping applications to rule out software issues, and in both cases, the user in Grant County was incorrectly mapped outside of the building over many attempts. However, another user was mapped correctly inside JPL (Jet Propulsion Laboratory) Pasadena, California. In all cases of erroneous GPS data, the user was either inside a building or next to a large wall. An AI agent that relies solely on GPS data is rendered useless in degraded GPS environments, such as inside buildings. Therefore, the onboard agent must be replaced by a completely different agent that can estimate their position within a building using other data from accelerometers or cell tower triangulation[15]. Furthermore, constant changes to configuration that enable or disable software features introduce the risk of human errors. The better solution is to have AI agents deployed when needed and manage their own lifecycle. The AI agent's lifecycle could be managed by external sensor data. In the case of an AI agent that relies on GPS, the agent can detect constant erroneous or noisy data and pause its execution. Then the agent can request another localization agent that is not reliant on GPS, rather relies on the smart phones accelerometer and gyroscope to perform Pedestrian Dead Reckoning (PDR)[16].

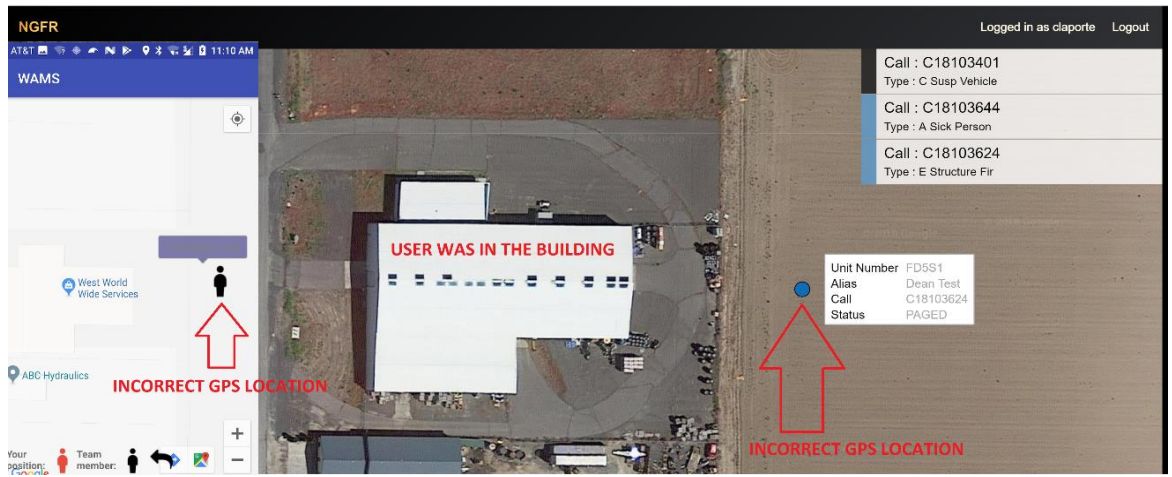

**Figure 2. User is mapped incorrectly via two independent software using the same GPS beacon.**

## Image Classification

To reduce medication errors and patient harm, paramedics attempt to adhere to the "five rights": the right patient, the right drug, the right dose, the right route, and the right time[17]. In emergency situations where there are multiple patients and high levels of stress, paramedics may perform one of the "five rights" incorrectly. Therefore, paramedics from Hastings County Canada, requested an AI assistant that could confirm that the "five rights" were satisfied. The authors integrated an AI model that was developed by another JPL team via TCP/IP sockets with the combination of Message Queuing Telemetry Transport (MQTT) to allow image classification, optical character recognition (OCR), and natural language processing (NLP). Since facial recognition and identification were already extensively studied and available commercially, we focused on 4 of the "five rights", which are the right drug, the right dose, the right route, and the right time. To determine the "right drug" the paramedic had to utter the phrases "Ok Audrey, take a picture" which alerted on board AI via NLP to open up the onboard camera application. The paramedic then could tap the screen or utter the phrase "Ok" to send the desired picture over the TCP/IP to the backend AI for classification. The backend AI would then perform image processing, and Optical Character Recognition (OCR) on the drug ampule label and determine the drug type. That drug title was returned via MQTT text and uttered back to the paramedic, such as "nitroglycerin". The challenges with this approach were the constant request and response between the mobile device and the backend AI. If the mobile device had fast and consistent connectivity, then all went well. However, due to poor cell tower signals more than 50% of the classifications required multiple attempts, never returned, or returned in unacceptable timeframes. After analyzing network logs, these issues were attributed to degraded cellular signals, while the experiments were performed in moving ambulances or rural areas. If the mobile application lost the entire network, then the application also lost all backend AI capabilities. One solution was to develop and deploy 2 models, one complex model such as Deep Neural Network (DNN) and another simpler neural network architecture such as a Multilayer Perceptron (MLP) to the mobile device. The problem with this approach is that simpler MLP models have less accuracy versus DNNs and are not suitable for applications that may affect a patient's wellbeing. Another unresolved issue is the need to manage a suite of models, which lends itself to a bloated monolithic application. A more sensible architecture would request and deploy the appropriate AI agent to the paramedic's mobile device specific to the emergency. There may only be the need for 1 or 2 network round trips, and AI agent deployment happens before paramedics are deployed to the field, which removes the risk of losing AI services in the field. Furthermore, entire models and corresponding software are encapsulated in a single agent which makes updating analytics software easier. In addition, containerization technologies such as Docker streamline the bundling of application code, model code, and other libraries[18]. Ultimately, developers can deploy minimal AI agent with only a minimal required software. The AI agent will remain dormant and does not consume any precious resources on edge devices until it is requested.

SOFTWARE CHALLENGES

The communication architecture described in **Figure 2** represent an oversimplified topology with the sole purpose of depicting network relationships between nodes. **Figure 3** represents a real-world tower topology that gathered a 1-hour recording session via JPL proprietary software. A user with a smartphone walked through a densely populated part of Los Angeles and gathered cell tower information. The black arrow depicts the path that the user walked, and the red and blue dots are actual cell tower positions. In **Figure 3** one can see that in the real-world cellular networks are a complex mix of 2G, 3G, 4G, and newer 5G towers.

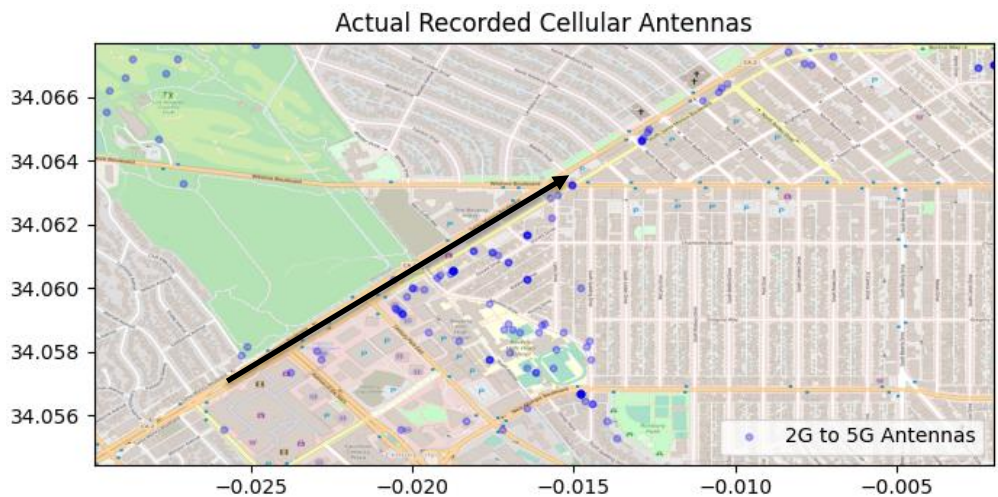

**Figure 3. Data Capture of Nearby Cellular Antennas in Order to Study Actual Urban Topology.**

Analyzing the raw data further shows fluctuating signal-to-noise ratios, bandwidth, and other performance metrics. The deployment of AI agents to edge devices requires consideration of current network conditions. In an emergency, AI agents may be deployed ahead of first responders to actual cell towers or 5G nodes to assess the network quality. AI agents may be deployed to smart cameras to gather intelligence ahead of first responders, to reduce the risks to emergency teams. Hence, only based on current conditions can an agent management software bundle pre-trained models with agents based on challenges that are unique to the specific emergency. If the initial emergency call reports a fire, then simultaneously onsite AI agents may detect hazardous chemicals, which alerts hazmat teams. Onsite network conditions may also dictate types and the configurations of AI agents. Perhaps the responder is deployed to a limited 4G or 5G environment, therefore AI agents with a deep neural model may be too network-intensive and the agent management software may bundle agents with simpler models, such as logistic regression that may rely on simpler data. Finally, the entire theme of the proposed second architecture is the right AI, deployed at the right time.

CONCLUSION AND FURTHER WORK

To assess the validity of the proposed architecture as depicted in **Figure 2**, we must test AI agents in 5G environments with SBA functionality. JPL is planning to set up simulated 5G test beds. The authors plan to examine many topologies, protocols, and network conditions that will help confirm or reject their hypothesis of the advantages of AI agents in 5G networks.

## ACKNOWLEDGMENTS

This work was done as a private venture and not in the author's capacity as an employee of the Jet Propulsion Laboratory, California Institute of Technology

The authors would like to thank David R Hanks JPL's Deep Learning Technology Group section supervisor for continued support.

**Aryan Naim** is a Senior Software Engineer in the Mission Control Systems Deep Learning Group at Jet Propulsion Laboratory, California Institute of Technology. aryan.e.naim@jpl.nasa.gov.

**Ryan Alimo,** received the Ph.D. degree in computational science from UC San Diego, CA, USA, in 2017 followed by a Postdoctoral Scholar with the Center for Autonomous Systems and Technology at California Institute of Technology. He is currently a machine learning scientist with the Deep Learning Technology Group at Jet

Propulsion Laboratory, California Institute of Technology. sralimo@jpl.nasa.gov.

**Jay Braun,** is Project Data Systems Engineer with Deep Space Network, Tracking, Telemetry, and Command at Jet Propulsion Laboratory, California Institute of Technology. Jay.E.Braun@jpl.nasa.gov.